# Deep Learning based Topic Analysis on Financial Emerging Event Tweets

Shaan Aryaman
School of Computer Science and Engineering
Nanyang Technological University
shaa0002@e.ntu.edu.sg

Nguwi Yok Yen
Nanyang Business School
Nanyang Technological Univeristy
yokyen@ntu.edu.sg

*Abstract* – Financial analyses of stock markets rely heavily on quantitative approaches in an attempt to predict subsequent or market movements based on historical prices and other measurable metrics. These quantitative analyses might have missed out on un-quantifiable aspects like sentiment and speculation that also impact the market. Analyzing vast amounts of qualitative text data to understand public opinion on social media platform is one approach to address this gap. This work carried out topic analysis on 28264 financial tweets [1] via clustering to discover emerging events in the stock market. Three main topics were discovered to be discussed frequently within the period. First, the financial ratio EPS is a measure that has been discussed frequently by investors. Secondly, short selling of shares were discussed heavily, it was often mentioned together with Morgan Stanley. Thirdly, oil and energy sectors were often discussed together with policy. These tweets were semantically clustered by a method consisting of word2vec algorithm to obtain word embeddings that map words to vectors. Semantic word clusters were then formed. Each tweet was then vectorized using the Term Frequency-Inverse Document Frequency (TF-IDF) values of the words it consisted of and based on which clusters its words were in. Tweet vectors were then converted to compressed representations by training a deep-autoencoder. K-means clusters were then formed. This method reduces dimensionality and produces dense vectors, in contrast to the usual Vector Space Model. Topic modelling with Latent Dirichlet Allocation (LDA) and top frequent words were used to analyze clusters and reveal emerging events.

## I. INTRODUCTION

Stock market trends is a variable with unknown number of predictors. One of the highly relevant predictor is investors' sentiments and speculations. Investors' sentiments about the market has great impact on both company's stock and the market as a whole. Quantitative methods to predict stock market trends usually do not pick up on emerging events or public speculation. This is where social media and especially microblogging sites like Twitter can help to provide data for a systemic analysis on public speculation. Twitter provides a channel to hear voice from the common public and amplifies the reporting and opinions of think tanks. Since tweets are limited to 280 characters, they are usually concise and provides good summary of the information by itself.

Two types of tweet data can be analysed to study stock markets trends. One, stock-specific tweets in a given timeframe. Such tweets could be from any sources of common public regardless of their influence (number of followers, likes and retweets). Or tweets can specifically be mined from influential and credible sources. Oussalah et. al. [2] takes this kind of approach by using tweets from US foreign policy, relevant think tanks and large oil companies and associations to predict oil prices. We took a similar approach. We used 28264 tweets by David Wallach on 'Financial Tweets' [1]. This dataset provides tweets on major exchanges. Information from NASDAQ, NYSE and SNP were scraped from influencers and think-tanks like Morgan Stanley, Reuters etc.

In order to identify emerging events, we have clustered the aforementioned tweets into semantic clusters. Usually for clustering textual data, text first needs to be appropriately mapped to a vector space. That is, each text document needs to be vectorized and then the vectors are subsequently clustered using clustering algorithms such as K-means. Usually vectorization techniques use bag of words or term frequency-inverse document frequency (TF-IDF) values to create vectors. Such vectors face issues of high dimensionality and sparsity. This high dimensionality may not lead to good clustering [3]. Our method uses the word2vec algorithm [4] that produces word embeddings for each word and hold significance of the semantic relation between words. These word-embeddings are clustered using K-means. Next, tweets are vectorized using contexture information like it is residing on which surrounding words within the TF-IDF values. Lastly, these tweet vector representations are further transformed to compressed representations using a deep-autoencoder. This reduces their dimensionality and sparsity. Finally, these compressed representations are clustered using K-means and topic modelling is done on these clusters.

This paper is organized as into 4 sections. Section II outlines related works that this paper builds upon and introduces imperative background concepts, Section III describes the methodology used in this work as well as experimental results. Lastly, section IV concludes the paper.

## II. RELATED WORKS AND BACKGROUND

This work builds on the approach of [5]. Fraj et. al. [5] proposed a novel approach to cluster tweets. In the approach all words in the corpus were used to train a word2vec model to obtain word-embeddings. These embeddings were clustered to form word clusters. Subsequently each tweet is vectorized using a formula that makes use of the TF-IDF values of words and the word clusters. This formula will be described in detail. Finally, these tweet vectors were clustered using K-means. We proposed an additional step on top of this approach to train a deep autoencoder on the tweet vectors and obtain compressed/encoded representation of tweet vectors before final clustering to further reduce dimensionality and sparsity.

Autoencoders have been described as useful models to obtain meaningful, compressed representation of data for clustering in [6]. Min et. al. [6] illustrated the use of autoencoders in Deep Embedded Clustering method (a deep learning-based clustering approach) wherein features train an autoencoder and the decoder part were dropped. The encoder network was then used as input for the clustering module.

This work aims to further enhance the approach was in [5] on the use of autoencoders and apply topic modelling using Latent Dirichlet Allocation (LDA) and topic frequent words to better analyze tweet clusters.

TF-IDF is a product between term frequency (TF) and inverse document frequency (IDF). TF measures how frequently a term occurs in a document. IDF gauges the importance of a term in the corpus of documents at large. Frequently occurring terms like 'a', 'an' may have high TF for each document but low IDF since they do not add much value to an individual document.

$$TF = \frac{frequency\ of\ term\ in\ document}{no.of\ terms\ in\ document} \qquad (1)$$

$$IDF = In\left(\frac{No\ of\ Documents}{No\ of\ Documents\ with\ term}\right) \qquad (2)$$

Word2vec is a set of neural networks that create word embeddings given a certain corpus of documents. There are primarily two classes of word2vec: continuous bag of words (CBOW) and skip-gram model [4]. We used the skip-gram model. This model, when given a word, provides the probability of the nearby words. Lower cosine distance between two-word embeddings corresponds to closer semantical relationship between the words and vice-versa.

K-means is a partitional clustering algorithm. In k-means a cluster can be characterized by one point known as the 'mean' or centroid. It follows an iterative process where every document is assigned to a cluster based on a distance metric. This is followed by revision of the centroid point by calculating the 'mean' of each cluster. The number of clusters to be formed must be specified.

Autoencoders are neural networks that perform lossy compression on data (they are not used for data compression) [7]. They compress the input data using an encoder function and reconstruct it using a decoder function. The distance function between the original data and the decompressed data is minimized. We used a deep autoencoder to obtain compressed representation of vectorized tweets as input for the final clustering.

Distance Metrics are functions that depict a distance between elements of a dataset. Euclidean and Cosine distance metrics were used in this work for distance calculation. Euclidean distance is the shortest distance between two points. Cosine similarity is the inner product between two normalized vectors. Given two vectors X and Y, cosine similarity is found using the following formulae.

$$Cosine - similarity = \frac{X.Y}{\|X\|\|Y\|} \qquad (3)$$

$$1 - cosine - similarity = cosine - distance \qquad (4)$$

Cosine similarity between two-word embeddings, produced by word2vec, of two semantically similar words is greater (or cosine distance between them is lesser) and vice-versa. At first, we wanted to transform the word vectors in a way such that cosine and Euclidean distance would be directly correlated. We could then apply many more clustering algorithms on them using Euclidean distance instead of cosine distance. We thought of normalizing the vectors, since cosine similarity only takes direction and not magnitude into account. However, this does not hold. This is illustrated below by two vectors, X and Y, that are normalized (have a magnitude of one):

$$\|X - Y\|^2 = (X - Y)^T(X - Y) = \|X\|^2 + \|Y\|^2 - 2X^TY \qquad (5)$$

$$\|X - Y\|^2 = 2 - 2\cos\_sim(X, Y) \qquad (6)$$

$$\therefore Cosine - distance^2 = 2 \times cosine - distance \qquad (7)$$

From (6) it is clear that the square of cosine distance between X and Y is directly proportional to cosine similarity between X and Y. This means that clustering of word embeddings would have to be done using cosine distance.

LDA [8] is an algorithm that aides in understanding topics in topic modelling, to identify the topics in a collection of documents. The number of topics is set arbitrarily, and each topic is represented by a set of words. Each word in the vocabulary is assigned to a topic. Essentially documents are distributed over topics and topics are distributed over words using Dirichlet distributions. LDA is based on a bag of words document, it assumes that words alone dictate the topics present in a document regardless of the syntactic and grammatic structure.

A recent work on LDA includes [9], which applies LDA to health-related tweets to identify topics present in them. In this work LDA has been used as an unsupervised model to help labeling unlabeled data which was later used to train a supervised model.

## III. METHODOLOGY

This work extends from the work depicted in [5]. Seven main steps were involved in the processing. The flow diagram is illustrated in Figure 1. The tweets data first goes through pre-processing prior to word2vec training. Followed by work embedding clustering, construction of tweet vector, training of auto-encoding to finally arrive at clusters suitable for topic modelling. The whole process mainly focuses on representation of shorts texts, tweets in this case, as dense vectors with low dimensionality.

The tweets used were taken from David Wallach's dataset and are not aimed at any specific company or stock. The method can also be generalized for dataset of tweets relates to one major event or stock.

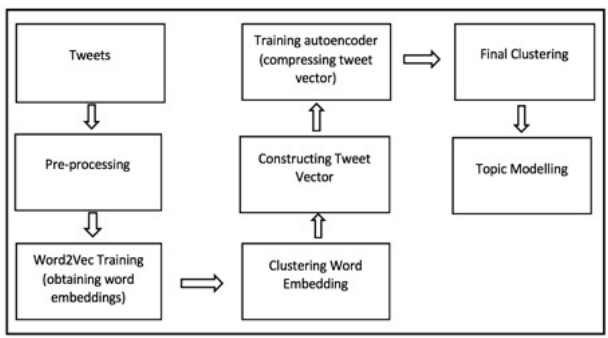

Figure 1. Methodology Flow

### A. PRE-PROCESSING

Since tweets can be noisy text data, it is important to pre-process them. We first removed punctuation, weblinks, mentions of usernames, removed all hashtags, removed stop words (high frequency words like 'a', 'the'), removed the 'RT' characters (specify whether the tweet was retweeted). The tweets were then converted to lower case and finally tokenized.

### B. OBTAINING WORD EMBEDDINGS

The processed, tokenized tweets were used to train a skip-gram word2vec model using the Gensim library's implementation of word2vec. We used negative sampling [10] to train our word2vec model. We set the negative sampling parameter to '10' which refers to the number of negative words that need to be drawn. The model produced word vectors with the dimension 300 for each word in the corpus.

### C. CLUSTERING WORD EMBEDDINGS

By clustering word embeddings, using cosine distance as a distance metric, semantic cluster of words would be obtained. In order to do so, we tried to cluster them on a range of clustering algorithms so as to see which algorithm performed the best.

Agglomerative Hierarchical is a type of Hierarchical Clustering Algorithm. This was the first algorithm we tried. We used Scipy library's implementation which allows the distance metric to be set to cosine distance. In order to select the number of clusters to be produced we needed to view the dendrogram (a diagram that records all sequences of mergers). Usually the number of clusters chosen are where there is maximum distance between all clusters on viewing the dendrogram. However, in the dendrogram it was not clear as to how many clusters should be formed. Either too few clusters would be formed, or certain clusters would have too many words assigned.

Density-based spatial clustering of applications with noise (DBSCAN) was the next algorithm we tried using with cosine distance as the distance metric. Density based clustering is useful when there are outliers in the data and clusters to be formed are not spherical and do not have equal density within a cluster. Hence, it is preferred to use DBSCAN when there are clusters with arbitrary shapes. In the case of the word embeddings that were obtained there was one big cluster formed with most vectors and just one or two other clusters with negligible number of vectors. This could be because the data does not have any significant outliers and there is uniform density throughout the vectors in the vector space. It could also indicate that the vectors were present in a somewhat spherical shape within the vector space.

K-means was the algorithm that was eventually used. It was also used in [5]. We clustered the word embeddings using NLTK's KMeansClusterer's implementation since it allows for k-means implementation with cosine distance as the distance metric.

### D. CONSTRUCTING TWEET VECTORS

Vectors to represent tweets were constructed from the semantic word clusters. These vectors have the same dimensionality as the number of word clusters, in our case there were 200 clusters. Each word cluster is given a specific weight. For example, the zeroth index in a tweet vector is assigned the weight of the zeroth word cluster in the tweet depending on which words are in the tweet also in the zeroth cluster.

$$w_{ab} = \frac{tf_{ab} \times idf_a}{\sum_{w \in T_b} tf_{ab} \times idf_a} \quad (8)$$

We used the formula in (8) from [5]. The equation describes the weight assigned to word cluster 'a' in tweet 'b'. The weight is the product between the sum of all TF values of all words present in cluster 'a' and tweet 'b' and the sum of IDF values of all words in the word cluster (regardless whether they are present in the tweet). The denominator is used to normalize the vector. The denominator is the same for all weights in the tweet vector.

### E. COMPRESSING TWEET VECTORS

A deep autoencoder was trained over the tweet vectors. The tweet vectors, with a dimensionality of 200, were compressed to vectors of a dimensionality of 20. Layers were stacked to form both the encoder and decoder part of the neural network where the encoder layers were successively reduced in output size from 200 to 20 and the decoder layer were successively increased in their output values from 20 to 200. The activation function in all layers was tanh except for the last layer where it was sigmoid. The optimizer was adadelta and the loss was binary-cross entropy. Binary cross entropy was used since the tweet vectors were normalized and [7] had also used binary cross entropy to compress normalized vectors. A loss of 0.0291 was achieved. After training the autoencoder, the decoder layers were removed, and the tweet vectors were input into the model to produce their corresponding compressed vectors. The compressed vectors are an alternate representation of tweet vectors while having less sparsity. This is evident from the fact that the model is able to accurately reproduce the original tweet vector from their compressed versions with a low loss of 0.0291. This greatly reduces the dimensionality of the data. With larger corpuses there could be many more than 200-word clusters formed. In such scenarios, the tweet vectors would also have very high dimensionality. The use of autoencoder could ensure that alternative representations that have lower dimension and denser vectors can be used for final clustering.

### F. FINAL CLUSTERING

The compressed vectors are clustered using k-means from sklearn library's implementation using Euclidean distance metric. There were 200 clusters formed.

### G. TOPICS MODELLING

LDA and the top 10 frequent words were found for each cluster. Five topics with five characteristic words were found for LDA as illustrated in the following section.

### H. RESULTS

The tweets were clustered into 200 clusters. Subsequent topic modelling reveals emerging events, speculation made by the public and possible analysis of upcoming trends in the stock market at large. LDA was applied to each of the clusters to produce five topics per cluster. The top five words characteristic for each topic is displayed with the word's weightage in the topic. The top ten most frequent words were also shown. The top results for cluster numbers 18, 24, 54 and 62 are shown below.

The first illustrative cluster obtained is Cluster 18 as shown in Table 1. The numbers beside each word denotes the corresponding weights. Each topic can be knitted together from the 5 keywords in each topic. For example, the first discussed topic is talking about the billion dollars of quarterly sales are expected to be announced in the respective period. It centers on the themes related to quarterly sales, results of companies. It also discussing about the earning per share, abbreviated as 'eps', of certain shares. Quarterly earning is one key figure that is frequently discussed. The top 10 frequently occurred words in this cluster are shown at the bottom of the table.

Table 1. Results of Cluster 18

| Topic No. | Cluster 18 | | |
|---|---|---|---|
| 0 | 0.080*sales | 0.080*billion | 0.075*quarterly |
| | 0.042*expect | 0.038*announce | |
| 1 | 0.068*q | 0.068*consensus | 0.066*sales |
| | 0.041*beats | 0.038*misses | |
| 2 | 0.059*earnings | 0.042*eps | 0.038*quarterly |
| | 0.036*results | 0.023*set | |
| 3 | 0.053*post | 0.047*quarterly | 0.045*earnings |
| | 0.032*inc | 0.029*share | |
| 4 | 0.068*eps | 0.063*expected | 0.044*company |
| | 0.037*quarter | 0.026*co | |
| **Top 10 frequent words:** downgraded, quarterly, amps, upgraded, sales, sdd, hold, code, ampd, sd. | | | |

Cluster 24 is shown in Table 2. It is on recurring theme of short selling shares. From the topics identified by LDA, there have been multiple mentions of the short interest, which is the number of shares that have been sold short but not yet being recovered. Morgan Stanley, a rating agency, has been identified as one of the most frequent words. The discussion on Ethereum, a cryptocurrency platform, was discovered in this cluster, as an alternative option after show selling.

Table 2. Results of Cluster 24

| Topic No. | Cluster 24 | | |
|---|---|---|---|
| 0 | 0.076*short | 0.060*volume | 0.040*interest |
|   | 0.037*sale | 0.023*unusually | |
| 1 | 0.095*short | 0.068*interest | 0.055*inc |
|   | 0.024*sale | 0.023*sees | |
| 2 | 0.182*short | 0.099*interest | 0.084*sale |
|   | 0.071*volume | 0.010*vol | |
| 3 | 0.063*inc | 0.051*cons | 0.051*pros |
|   | 0.051*fundamental | 0.051*explore | |
| 4 | 0.043*buy | 0.034*ethereum | 0.034*basechain |
|   | 0.034*mns | 0.034*aid | |
| **Top 10 frequent words:** volume, code, ampd, sd, interest, rating, morgan, stanley, price, short, target | | | |

Table 3. Results of Cluster 54

| Topic No. | Cluster 54 | | |
|---|---|---|---|
| 0 | 0.064*filing | 0.049*new | 0.037*insider |
|   | 0.034*sec | 0.031*form | |
| 1 | 0.067*stock | 0.065*inc | 0.064*filing |
|   | 0.064*sec | 0.040*alerts | |
| 2 | 0.052*international | 0.048*flavors | 0.048*fragrances |
|   | 0.046*shares | 0.044*am | |
| 3 | 0.072*insider | 0.069*stock | 0.054*sells |
|   | 0.047*inc | 0.039*shares | |
| 4 | 0.064*inc | 0.057*sells | 0.052*stock |
|   | 0.050*filing | 0.047*sec | |
| **Top 10 frequent words:** sd, sec, declined, llc, filing, registration, stock, inc, expiration, maxpain. | | | |

Cluster 54 is shown in Table 3. It is characterized by the topic of insider trading. Topic 1 has mentioned the word 'alerts' along with 'sec', which is referring to the US Securities and Exchange Commission, a regulatory body. Topic 3 mentions the words 'insider', stock' and 'sells'.

Cluster 62 is shown in Table 4. It is characterized by tweets referring to the oil and energy sectors. Its topics highlight words such as 'oil', 'energy', 'US' (as United States' foreign policy affects and is affected by oil prices greatly). A company, forming the word 'petroteqenergy' (a Canadian company innovating in heavy oil processing and extraction technologies), was also mentioned frequently. There were some positive sentiment being picked up within this cluster, with the word of bullish used on telcoin, another cryptocurrency based on Ethereum blockchain.

Table 4. Results of Cluster 62

| Topic No. | Cluster 62 | | |
|---|---|---|---|
| 0 | 0.028*oil | 0.022*inc | 0.022*libya |
|   | 0.013*eog | 0.013*force | |
| 1 | 0.013*aprn | 0.013*top | 0.013*mfgp |
|   | 0.013*uxin | 0.013*lmfa | |
| 2 | 0.023*price | 0.023*expect | 0.021*bull |
|   | 0.021*telcoin | 0.021*investinhd | |
| 3 | 0.028*energy | 0.028*oil | 0.026*jul |
|   | 0.025*pqeff | 0.022*petroteqenergy | |
| 4 | 0.074*trade | 0.073*us | 0.071*exchanges |
|   | 0.071*dfs | 0.071*find | |
| **Top 10 frequent words:** sdd, rose, code, holding, ampd, sd, sec, declined, llc, filing. | | | |

The topics identified in clusters can be indicative of general opinions from the public, think-tanks and news media. A further extension of this work could be the set-up of a dashboard system with more in-depth analysis of the various topics being discussed in tweets, the number of tweets discussing them and the influence (number of followers, likes) etc. Such text analysis can complement existing quantitative techniques. It can indicate possible reasons behind a particular market behavior and point to emerging events that have caught public attention, regarding any incident, product or stock. This may give clues to future trends and behaviors.

## IV CONCLUSION

We clustered financial tweets into semantic clusters to identify emerging trends. The clustering method we used focus on reducing dimensionality and sparsity of tweet representations (vectors) in order to better clustering performance. Topic modelling was used to better understand the clusters. One weakness is the fact that certain clusters might not be meaningful, and our method does not focus on identifying anomalies that might come in the form of tweets focused on advertising or tweets from bots. This could be rectified by further study and use other clustering algorithms or techniques that can also identify anomalies.